\documentclass[runningheads]{llncs}

\usepackage{eccv}

\usepackage{multirow}
\usepackage{color}
\usepackage{subcaption}
\usepackage{eccvabbrv}

\usepackage{graphicx}
\usepackage{booktabs}

\usepackage[accsupp]{axessibility}  

\usepackage{hyperref}

\usepackage{orcidlink}

\begin{document}

\title{Gr-IoU: Ground-Intersection over Union for Robust Multi-Object Tracking with 3D Geometric Constraints}

\titlerunning{Gr-IoU: Ground-Intersection over Union}

\author{Keisuke Toida\inst{1}\orcidlink{0009-0006-4873-3651} \and
Naoki Kato\inst{2}\orcidlink{0009-0004-3815-0829} \and
Osamu Segawa\inst{2}\orcidlink{0009-0000-2469-6098} \and
Takeshi Nakamura\inst{2}\orcidlink{0009-0001-4991-3383} \and
Kazuhiro Hotta\inst{1}\orcidlink{0000-0002-5675-8713}
}

\authorrunning{K.Toida et al.}

\institute{Meijo University, 1-501 Shiogamaguchi, Tempaku-ku, Nagoya 468-8502, Japan \and
Chubu Electric Power Co., Inc., 1-1 Higashishin-cho, Higashi-ku, Nagoya 461-8680, Japan\\
}
\maketitle

\begin{abstract}
We propose a Ground IoU (Gr-IoU) to address the data association problem in multi-object tracking.
When tracking objects detected by a camera, it often occurs that the same object is assigned different IDs in consecutive frames, especially when objects are close to each other or overlapping.
To address this issue, we introduce Gr-IoU, which takes into account the 3D structure of the scene.
Gr-IoU transforms traditional bounding boxes from the image space to the ground plane using the vanishing point geometry.
The IoU calculated with these transformed bounding boxes is more sensitive to the front-to-back relationships of objects, thereby improving data association accuracy and reducing ID switches.
We evaluated our Gr-IoU method on the MOT17 and MOT20 datasets, which contain diverse tracking scenarios including crowded scenes and sequences with frequent occlusions.
Experimental results demonstrated that Gr-IoU outperforms conventional real-time methods without appearance features.

\end{abstract}

\section{Introduction}
\label{sec:intro}

Multi-Object Tracking (MOT) is a significant and extensively studied problem in computer vision.
The primary goal of MOT is to consistently track multiple objects in a video sequence, assigning unique IDs to each object and accurately tracking their movements across frames. 
However, MOT presents several technical challenges.
One of the major issues in MOT is data association errors.
When objects are close to each other or overlapping, the same object may be assigned different IDs in consecutive frames,  causing ID switches and reducing the tracking accuracy.

\begin{figure}[tb]
    \centering
    \begin{minipage}[b]{0.48\linewidth}
        \centering
        \includegraphics[height=3.5cm]{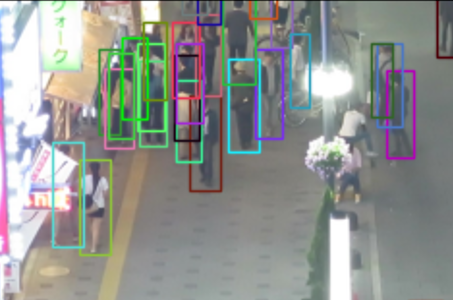}
        \subcaption{conventional b-boxes}
        \label{fig:bbox}
    \end{minipage}
    \begin{minipage}[b]{0.48\linewidth}
        \centering
        \includegraphics[height=3.5cm]{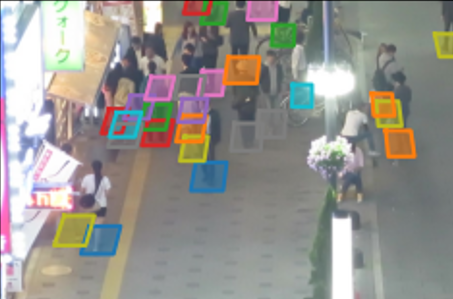}
        \subcaption{transformed b-boxes}
        \label{fig:bbox-gr}
    \end{minipage}
    \caption{Visualization of Gr-IoU b-boxes}
\end{figure}

In this work, we introduce new constraints to address these issues, focusing on objects that move on the ground plane (e.g., people and vehicles), which objects cannot overlap in 3D space by their physical nature.
We propose a novel matching method that incorporates the 3D structure of the scene.
Our proposed Gr-IoU (Ground-Intersection over Union) leverages vanishing point geometry to transform traditional bounding boxes from image space to the ground plane.
Transformed bounding boxes are visualized in \cref{fig:bbox-gr}.
The IoU computed with these transformed bounding boxes is more sensitive to the spatial relationships between objects, thereby improving data association accuracy and reducing ID switches.

We conducted experiments on both the MOT17\cite{milan2016mot16} and MOT20\cite{dendorfer2020mot20} datasets to evaluate the performance of our proposed Gr-IoU method.
Experimental results demonstrated that our approach outperforms conventional methods that do not utilize appearance features.
Specifically, our method shows significant improvements in reducing ID switches and increasing tracking accuracy across various scenarios, including crowded scenes and those with frequent occlusions.
These findings indicate that Gr-IoU which incorporates 3D geometric information, is effective for tracking multiple objects.

This paper is organized as follows. \Cref{sec:related} describes related works. In \cref{sec:method}, the details of the proposed method is explained. Experimental results are shown in \cref{sec:experiments}. Finally, we describe conclusions and future works in \cref{sec:conclusion}.

\section{Related Works}
\label{sec:related}

Data association is a critical component in MOT, involving the matching of detected objects across consecutive frames while maintaining consistent object identities.
Various methods\cite{bewley2016sort, wojke2017deepsort, chen2018motdt, Bergmann_2019tracktor, zhang2021fairmot, wang2019towardsjde, zhou2020tracking, zhang2022bytetrack} have been developed to address the challenges about data association.

The Kalman filter\cite{kalman1960new} is widely used for predicting the future positions of objects based on their previous states.
The Hungarian algorithm\cite{kuhn1955hungarian}, on the other hand, is employed to solve the assignment problem, matching predicted positions with detected objects.
Although they are effective in simple scenarios, these methods can struggle with occlusions and complex interactions between objects.

By the advent of deep learning, several learning-based approaches have been proposed to improve data association in MOT.
DeepSORT\cite{wojke2017deepsort}, for instance, extends the traditional SORT\cite{bewley2016sort} algorithm by integrating deep learning-based appearance features.
It employs a convolutional neural network (CNN) to extract appearance features of detected objects, which are then combined with motion information to enhance data association.
This method has demonstrated significant improvements in tracking performance, particularly in scenarios with visually similar objects.

Other deep learning-based methods, such as \cite{aharon2022bot, cao2023observation, du2023strongsort, wang2024smiletrack}, have also shown promising results in terms of tracking accuracy.
However, these approaches often come with a significant computational cost, limiting their applicability in real-time scenarios.
High computational requirements of deep neural networks, especially when processing high-resolution video streams, can lead to substantial latency, making them impractical for many real-world applications that demand real-time performance.

Our proposed Gr-IoU method aims to address the limitations of both traditional and deep learning-based approaches.
By incorporating 3D scene structure into the data association process, Gr-IoU improves the tracking accuracy without the computational overhead associated with deep learning methods.
This geometric approach allows for efficient processing, making it suitable for real-time applications while maintaining robust performance in complex scenes with frequent occlusions and dense object interactions.

Gr-IoU leverages vanishing point geometry to transform bounding boxes from image space to the ground plane, enabling more accurate spatial reasoning.
This approach not only improves data association accuracy but also maintains computational efficiency, striking a balance between performance and real-time applicability that is often challenging to achieve with deep learning-based methods.

\section{Proposed Method}
\label{sec:method}

\begin{figure}[tb]
  \centering
  \includegraphics[height=4.5cm]{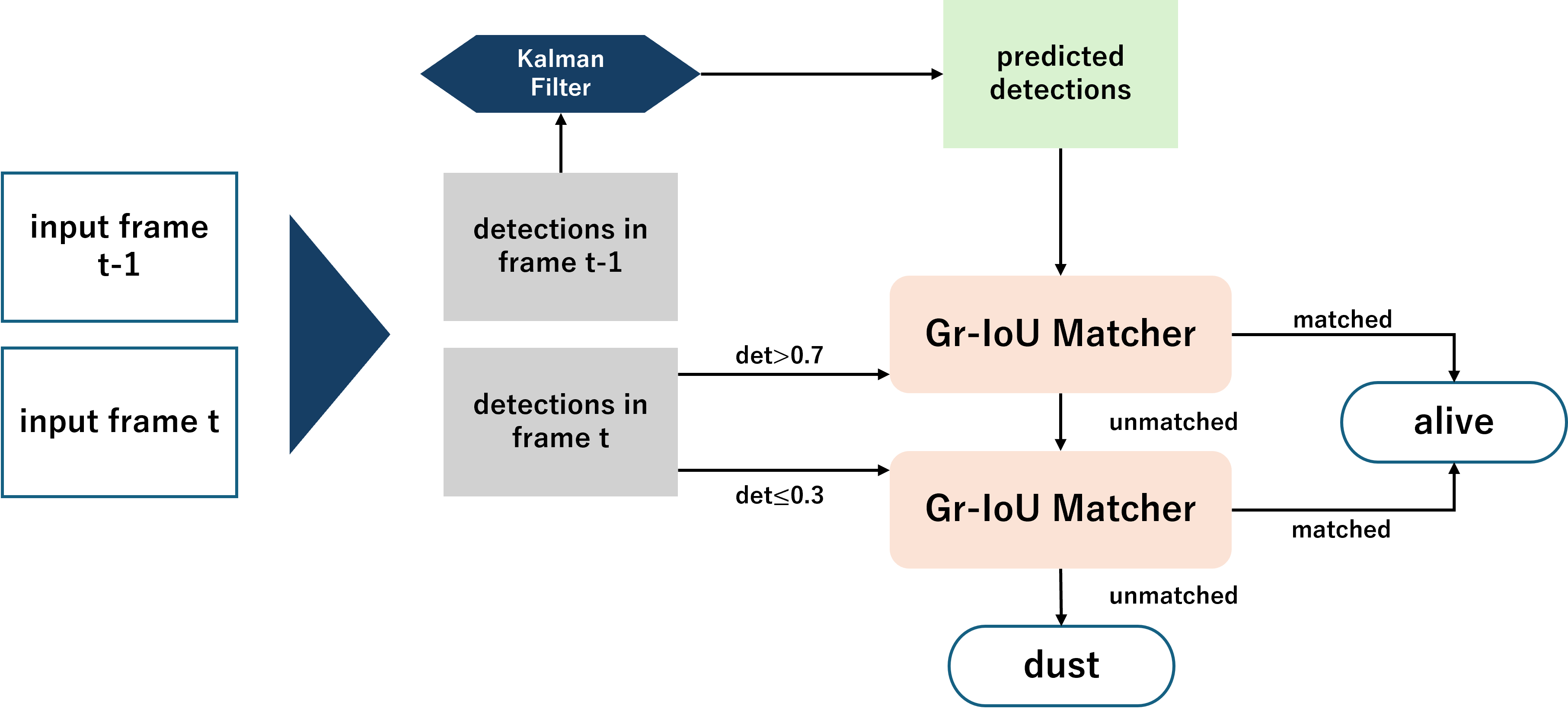}
  \caption{\textbf{Our tracking pipeline.}
  Our tracking pipeline follows the paradigm proposed in \cite{zhang2022bytetrack}, dividing the matching process based on the detection precision score (det).}
  \label{fig:pipeline}
\end{figure}

\Cref{fig:pipeline} illustrates the architecture of our proposed tracking method, which incorporates the Ground IoU (Gr-IoU) method.
It is designed to be able to assign correct IDs of close or overlapping objects without using visual features.
In \cref{sec:pipeline}, we provide the overview of pipeline for multi-object tracking, outlining the fundamental steps and components involved in the tracking process. 
\Cref{sec:gr-iou} introduces and explains in detail our novel Gr-IoU (Ground-Intersection over Union) algorithm.

\subsection{Pipeline}
\label{sec:pipeline}

We follows the tracking-by-detection paradigm.
Tracking-by-detection is a widely used framework in multi-object tracking (MOT).
This approach involves two main steps.
At first, objects are detected in each frame of a video sequence, and then linking these detections across consecutive frames to form consistent object trajectories.
In our tracking system, we first use an off-the-shelf object detector (e.g., YOLOX\cite{ge2021yolox}) to generate bounding boxes for objects in each frame.
Then, our proposed tracking system takes the detected bounding boxes as input and generates a continuous object trajectory as the final tracking result.

Our tracking system incorporates a Kalman filter\cite{kalman1960new} as a motion model to linearly predict the motion of objects.
The state variables of the Kalman filter are expressed by the following equation:
\begin{align}
    \textbf{x} = [x_c, y_c, h, a, \dot{x_c}, \dot{y_c}, \dot{h}, \dot{a}]
\end{align}
where $(x_c, y_c)$ are the center coordinates of the bounding box, $h$ is the height, $a$ is the aspect ratio, and $(\dot{x_c}, \dot{y_c}, \dot{h}, \dot{a})$ represent their respective velocities.

To improve data association accuracy, we utilize our proposed Gr-IoU method to calculate the cost matrix.
This geometric approach enhances the spatial reasoning capabilities of the tracker.
Details are given in \cref{sec:gr-iou}.
Finally, we employ the Hungarian algorithm to solve the assignment problem based on this cost matrix, effectively linking detections across frames.

\subsection{Ground-IoU}
\label{sec:gr-iou}

\begin{figure}[tb]
  \centering
  \includegraphics[height=4.5cm]{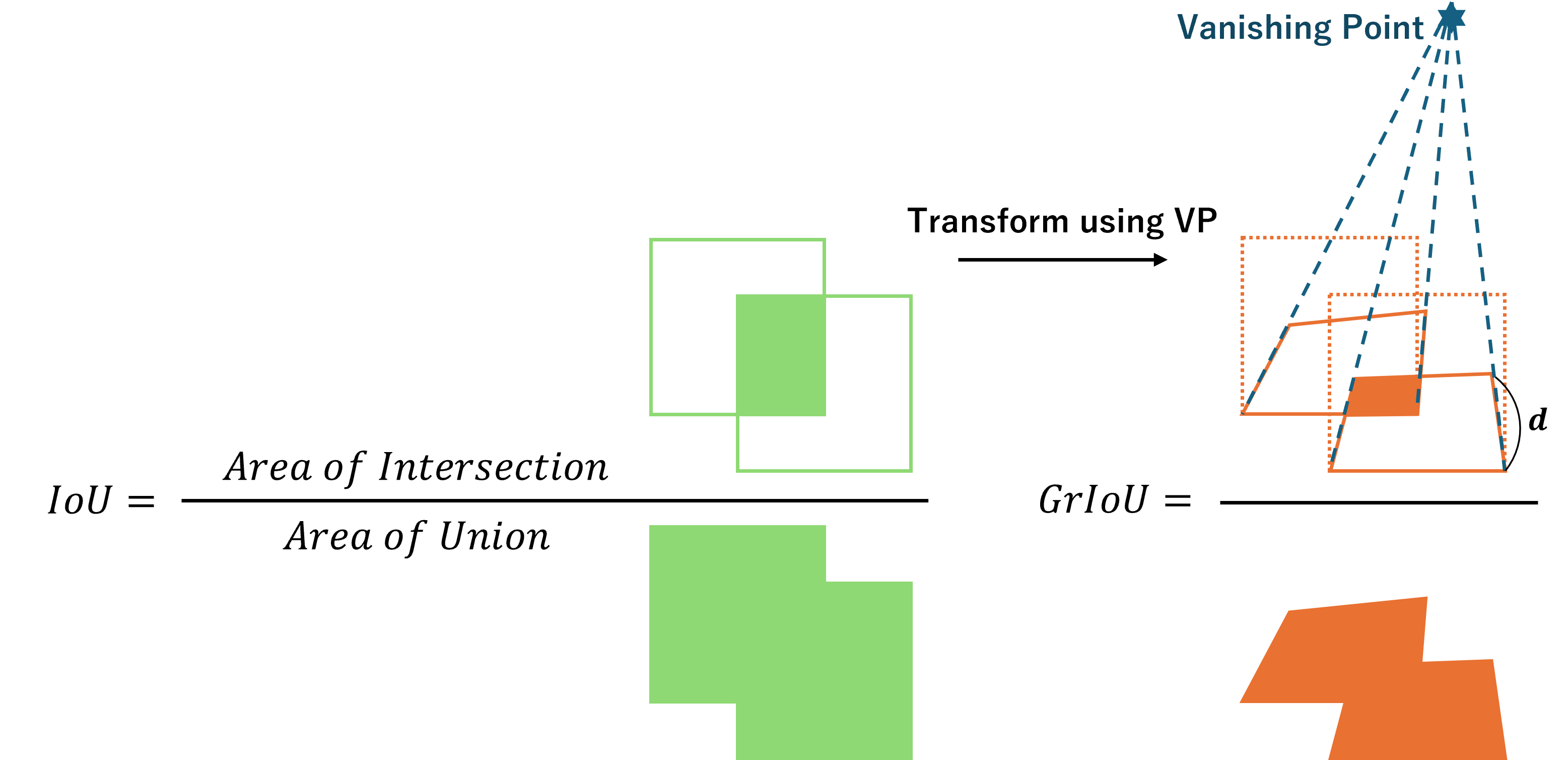}
  \caption{
  \textbf{Overview of Gr-IoU.}
  In conventional cost matrix calculations, standard IoU in camera space is used.
  In our method, we transform the coordinates of the detected bounding boxes by projecting them onto the ground plane.
  By calculating IoU using these transformed rectangles, we alleviate redundancy in the cost matrix and enable more efficient matching for close objects and occlusions.
  }
  \label{fig:griou}
\end{figure}

The main contribution proposed in this paper is Ground-Intersection over Union (Gr-IoU).
Gr-IoU addresses a key issue in traditional tracking methods: redundancy in the cost matrix caused by occlusion or when objects appear close together in 2D image space.
In traditional 2D approaches, when objects overlap or occlude each other, the IoU cost matrix often fails to accurately represent the real spatial relationships.
Gr-IoU significantly mitigates this issue by incorporating 3D geometric information by projecting bounding boxes onto the ground.
This projection is performed using the vanishing point estimated in the initial frame.
Gr-IoU is calculated using the rectangle formed by the four projected points, as shown in \cref{fig:griou}.

The left side in \cref{fig:griou} shows conventional method of IoU calculation, which uses bounding boxes coordinates in camera space.
In contrast, our proposed method shown on the right side in \cref{fig:griou} uses bounding boxes coordinates projected onto the ground plane to calculate IoU.
By using our method on ground plane, it is robust to close and occluded objects. 

Let the XY-coordinates of the four points of the bounding box in camera space obtained by the detector be denoted as $(i_{tl}, i_{tr}, i_{bl}, i_{br})$.
The XY-coordinates of these four points transformed to project onto the ground plane are represented as $(o_{tl}, o_{tr}, o_{bl}, o_{br})$.
Given the vanishing point coordinates $(x_{vp}, y_{vp})$, the transformation of each point can be expressed by \cref{eq:2}.
\begin{equation}
\label{eq:2}
    (o_{tl}, o_{tr}, o_{bl}, o_{br}) = (t_{tl}, t_{tr}, i_{bl}, i_{br})
\end{equation}
where transformed points $t_{tl}, t_{tr}$ are calculated as in \cref{eq:3}.
\begin{equation}
\label{eq:3}
    t(x_{t}, y_{t}) = (x_{i} + du_{x}, y_{i} + du_{y})
\end{equation}
where $i(x_i, y_i)$ is a bottom input point, $t(x_t, y_t)$ is a transformed point, $(u_x, u_y)$ is a unit vector in the direction of the vanishing point, and $d$ is a parameter controlling how far the new point is from the original point.

The unit vector $(u_x, u_y)$ is calculated as in \cref{eq:4}.
\begin{equation}
\label{eq:4}
    (u_x, u_y) = \left(\frac{x_{vp}-x_i}{\sqrt{(x_{vp}-x_i)^2+(y_{vp}-y_i)^2}}, \frac{y_{vp}-y_i}{\sqrt{(x_{vp}-x_i)^2+(y_{vp}-y_i)^2}}\right)
\end{equation}

This transformation projects the top points of the bounding box towards the vanishing point, creating a trapezoid shape that better represents the object's position on the ground plane.
The bottom points remain unchanged, anchoring the object to its original position in the image.
In our experiments, we set $d = 0.3h$ ($h$ is a height of bounding box) so that it is proportional to the size of the bounding box.
We also add a scaling buffer to make the cost matrix more responsive.

This projection helps resolve ambiguity in object associations and allows for a more accurate representation of spatial relationships even in complex scenarios.
As a result, Gr-IoU reduces the likelihood of tracking errors caused by misleading camera space similarity scores, especially for partially occluded or closely spaced objects.

\section{Experiments}
\label{sec:experiments}

This section describes our experimental setup, evaluation methodology, and results.
In \cref{sec:datasets}, We detail the datasets used in experiments and the specific conditions under which they were employed.
Then, in \cref{sec:metrics}, we explain the quantitative measures used to assess the performance of our Gr-IoU method and compare it with existing trackers.
Next, in \cref{sec:results}, we present a comparison between our proposed Gr-IoU method and baseline trackers.
This subsection includes a detailed analysis of the results, highlighting the strengths and limitations of our approach.

\subsection{Datasets}
\label{sec:datasets}

Our experiments were conducted on the MOT17\cite{milan2016mot16} and MOT20\cite{dendorfer2020mot20} training datasets with private detections by YOLOX\cite{ge2021yolox} ablation model from \cite{zhang2022bytetrack}.
These datasets provide diverse scenarios for evaluating multi-object tracking algorithms.
For the MOT17 dataset, we evaluate the performance of Gr-IoU using only static camera sequences and vanishing points estimated by ELSED\cite{suarez2022elsed} and RANSAC\cite{fischler1981random}.
We estimate the vanishing point in the first frame of each sequence and use it for all subsequent frames.

In contrast, for the MOT20 dataset, we adopt a simplified approach where the vanishing point is set to (image\_width/2, 0).
This decision was made because MOT20 contains sequences captured by non-static cameras, where per-frame vanishing point estimation can be computationally expensive and potentially unreliable.
By using a fixed vanishing point at the top center of the image, we maintain a reasonable approximation of perspective while avoiding the computational overhead of dynamic vanishing point estimation.
This adaptive approach to vanishing point determination allows our Gr-IoU method balancing accuracy and computational efficiency.

\subsection{Metrics}
\label{sec:metrics}

In this section, we describe the evaluation metrics used to evaluate the performance of our proposed Gr-IoU method on the MOT17 dataset.
The primary metric is Multi-Object Tracking Accuracy (MOTA)\cite{bernardin2008evaluating}, which provides an overall measure of tracking accuracy by combining false positives, false negatives, and identity switches, thereby penalizing both detection and association errors.
MOTA is a comprehensive metric that reflects the general performance of a tracking system, with higher scores indicating better tracking performance.

In addition to MOTA, we utilize other standard metrics commonly used in multi-object tracking evaluation, including Identity F1 Score (IDF1), Identity Precision (IDP), Identity Recall (IDR), Mostly Tracked (MT), Mostly Lost (ML), Partly Tracked (PT), and Identity Switches (IDsw).

\subsection{Results}
\label{sec:results}

\subsubsection{Comparison on MOT17 training dataset with private detections.}
\label{sec:mot17}

\Cref{tab:mot17} compares our Gr-IoU to mainstream MOT methods on the training sets in MOT17 dataset.
We compared our method with ByteTrack\cite{zhang2022bytetrack}, which is renowned for its high accuracy among tracking methods that do not utilize appearance features.

Gr-IoU achieves the highest overall MOTA (79.10\%) and IDF1 (80.81\%), indicating superior tracking accuracy and identity consistency compared to SORT and ByteTrack.
Notably, Gr-IoU also exhibits the highest IDP (87.38\%), highlighting its precision in identity matching.
Additionally, Gr-IoU shows a significant reduction in identity switches, further underscoring its effectiveness in maintaining consistent object identities over time.
These results confirm that incorporating 3D constraints through Gr-IoU enhances the overall performance of multi-object tracking.

\newcommand{\bhline}{\noalign{\hrule height 1.1pt}} 

\begin{table}[tb]
\caption{Comparison on MOT17 training dataset with private detections.}
\label{tab:mot17}
\centering
    \scalebox{0.9}{
    \begin{tabular}{c|c|cccccccc}
\bhline
&dataset&MOTA&IDF1&IDP&IDR&MT&PT&ML&IDsw    \\ \hline
\multirow{4}{*}{SORT}       &MOT17-04&	86.95\%&	90.08\%&	91.72\%&	88.50\%&	58&	9&	2&	34 \\
                            &MOT17-09&	79.54\%&	81.26\%&	91.44\%&	73.12\%&		15&	6&	1&	6  \\  
                            &MOT17-02&	51.08\%&	52.07\%&	63.80\%&	43.99\%&		14&	31&	8&	97 \\
                            &OVERALL&   76.78\%&	80.28\%&	85.84\%&	75.39\%&	87&	\textbf{46}&	\textbf{11}&	137  \\
\hline
\multirow{4}{*}{ByteTrack}  &MOT17-04 &87.87\%&   89.00\%&    89.55\%&    88.45\%&       61&  6&    1602&   26 \\
                            &MOT17-09 &83.15\%&	77.87\%&	85.15\%&	71.73\%&		16&	 5&	 1&		11 \\
                            &MOT17-02 &52.94\%&	55.51\%&	64.83\%&	48.53\%&		19&	26&	  8&	60 \\
                            &OVERALL	    &78.15\%&	79.97\%&	83.81\%&	\textbf{76.47}\%&	\textbf{96}&	37&	\textbf{11}&	97  \\
\hline
\multirow{4}{*}{Gr-IoU}     &MOT17-04	&88.83\%&	90.93\%&	93.12\%&	88.84\%&		60&	7&	2&	15   \\
                            &MOT17-09	&81.76\%&	72.74\%&	80.40\%&	66.41\%&	16&	5&	1&	10  \\
                            &MOT17-02	&54.52\%&	53.92\%&	69.08\%&	44.22\%&		14&	27&	12&	46 \\
                            &OVERALL	    &\textbf{79.10}\%&	\textbf{80.81}\%&	\textbf{87.38}\%&	75.16\%&90&	39&	15&	\textbf{71} \\                 
\bhline
    \end{tabular}
    }
\end{table}

\begin{table}[tb]
\caption{\textbf{Comparison on MOT20 training dataset with private detections.}
Gr-IoU\dag \,\, refers to a simplified version of Gr-IoU where the vanishing point is not estimated but is set to a fixed value.}
\label{tab:mot20}
\centering
    \scalebox{0.9}{
    \begin{tabular}{c|c|cccccccc}
\bhline
&dataset&MOTA&IDF1&IDP&IDR&MT&PT&ML&IDsw    \\ \hline
\multirow{5}{*}{ByteTrack}  &MOT20-02&	68.97\%&	62.34\%&	67.87\%&	57.65\%&		105&	64&	5&		329   \\
                            &MOT20-05&	61.83\%&	66.25\%&	68.96\%&	63.75\%&  324&	246&	87&		756    \\
                            &MOT20-03&	69.30\%&	76.60\%&	81.67\%&	72.13\%&  	260&	187&	70&		249    \\  
                            &MOT20-01&	67.76\%&	77.47\%&	81.77\%&	73.61\%&  	40&	23&	7&		38  \\
                            &OVERALL&	65.35\%&	69.11\%&	72.95\%&	65.65\%&   	729&	\textbf{520}&	\textbf{169}&		1372   \\
\hline
\multirow{5}{*}{Gr-IoU\dag}     &MOT20-02&	69.44\%&	64.42\%&	69.99\%&	59.67\%&  	107&	62&	5&		296   \\
                            &MOT20-05&	62.14\%&	67.46\%&	70.12\%&	64.98\%&  	330&	238&	89&		650    \\
                            &MOT20-03&	69.37\%&	77.63\%&	82.86\%&	73.01\%&  	255&	192&	70&		209    \\
                            &MOT20-01&	67.71\%&	77.08\%&	81.36\%&	73.23\%&  	40&	23&	7&		32  \\
                            &OVERALL&	\textbf{65.60}\%&	\textbf{70.35}\%&	\textbf{74.22}\%&	\textbf{66.86}\%& 	\textbf{732}&	515&	171&	\textbf{1187}  \\
                            \bhline
    \end{tabular}
    }
\end{table}

\subsubsection{Comparison in MOT20 training dataset with private detections.}
\label{sec:mot20}

\Cref{tab:mot20} illustrates the advantages of our proposed Gr-IoU of simplified version denoted as Gr-IoU\dag, when evaluated on the MOT20 training dataset with private detections.
Gr-IoU\dag \,\,achieves higher overall MOTA (65.60\%) and IDF1 (70.35\%), indicating improved tracking accuracy and identity consistency compared to ByteTrack.
Additionally, Gr-IoU\dag \,\,demonstrates superior IDP (74.22\%) and IDR (66.86\%), reflecting its effectiveness in identity matching and recall.
Notably, Gr-IoU\dag \,\,also reduces the number of identity switches (1,372 to 1,187), emphasizing its capability to maintain consistent object identities over time.
These results highlight that even the simplified version of Gr-IoU, with a fixed vanishing point, enhances the performance of multi-object tracking on the challenging MOT20 dataset.

\subsubsection{Ablation studies and other experiments}

We performed additional experiments on the MOT17 training dataset to evaluate the sensitivity of our method to the parameter $d$. \Cref{tab:d} shows the impact of different values of $d$ on MOTA and IDsw.
For $d$ ranging from $0.1h$ to $0.4h$, the MOTA remains relatively stable.
However, when $d$ exceeds $0.5h$, the MOTA begins to deteriorate, suggesting that larger values of $d$ lead to a decline in tracking performance.
These results suggest that this phenomenon occurs because a too large value of d causes the transformed bounding box to expand unnecessarily as shown in \Cref{fig:griou}, adversely affecting other bounding boxes that do not accurately represent the spatial relationships between objects.

\begin{figure}[tb]
    \centering
    \begin{minipage}[b]{0.48\linewidth}
        \centering
        \includegraphics[height=4.4cm]{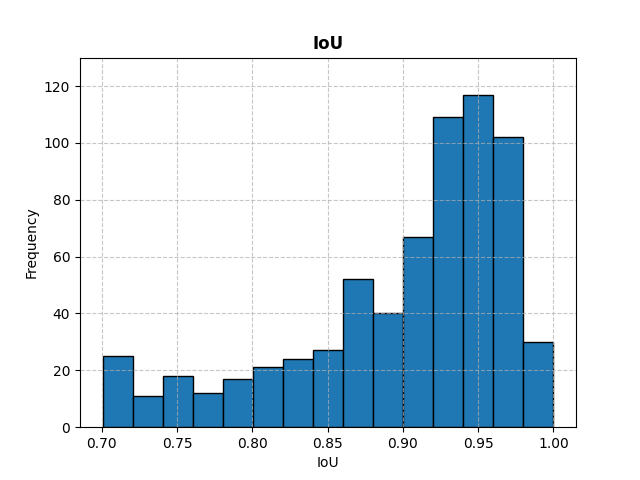}
        \subcaption{Conventional IoU}
        \label{fig:hista}
    \end{minipage}
    \begin{minipage}[b]{0.48\linewidth}
        \centering
        \includegraphics[height=4.4cm]{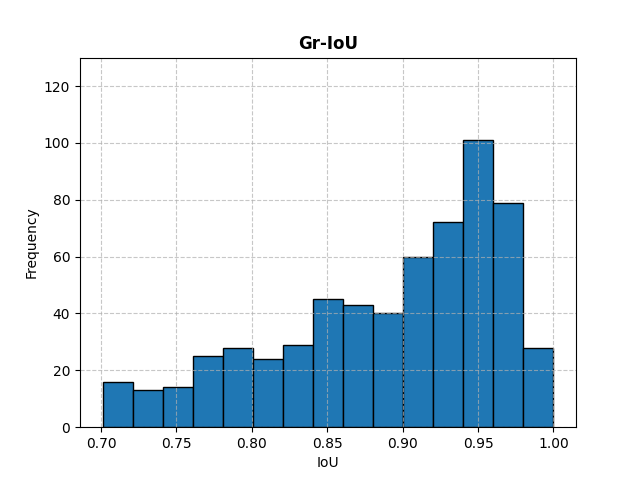}
        \subcaption{Gr-IoU}
        \label{fig:histb}
    \end{minipage}
    \caption{Histogram of IoU used to calculate the cost matrix}
    \label{fig:hist}
\end{figure}

\Cref{fig:hist} shows the IoU distribution used in calculating the cost matrix visualized as a histogram.
The vertical axis represents frequency, and the horizontal axis represents IoU value, focusing on the distribution where the IoU is close to 1.

\Cref{fig:hista} illustrates the distribution of conventional IoU values in image space, while \cref{fig:histb} depicts the distribution of Gr-IoU values.
These histograms demonstrate that the proposed Gr-IoU method mitigates the concentration of values near 1, resulting in a more uniform distribution across the range of IoU values.
The prevalence of cost matrix with numerous IoU values close to 1 adversely affects the optimization of the Hungarian algorithm\cite{kuhn1955hungarian}, which performs one-to-one data association.
This is because IoU values close to 1 are difficult to distinguish and introduce ambiguity in the association process.
The improved distribution of IoU values in the cost matrix achieved by the Gr-IoU method likely enhances the solvability of the assignment problem.
Consequently, this improvement contribute to a reduction in identity switches (IDsw).

\begin{figure}[tb]
  \centering
  \includegraphics[height=5.7cm]{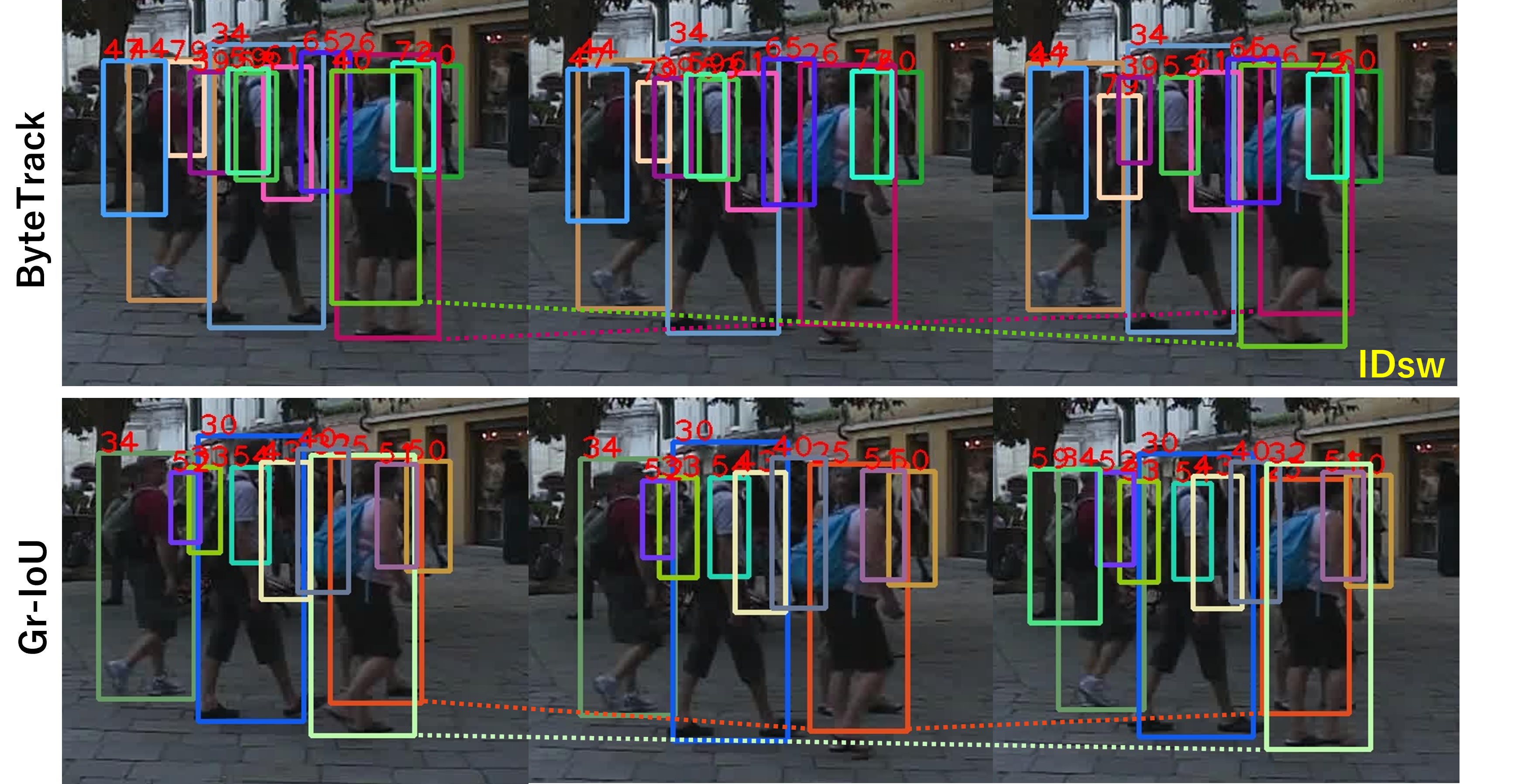}
  \caption{Example that ID swich was improved}
  \label{fig:idsw}
\end{figure}

\Cref{fig:idsw} shows a comparative analysis of Gr-IoU and ByteTrack with bounding boxes visualization per frame.
In results by the ByteTrack, we observe an ID switch in the frames immediately before and after a detection error.
In contrast, Gr-IoU improved ID switch and better tracking consistency across frames with detection uncertainty.

\begin{table}[tb]
    \caption{The impact of different values of $d$ on MOT17 training dataset}
    \label{tab:d}
    \centering
    \scalebox{0.9}{
    \begin{tabular}{c|cccccc}
            \bhline
            $d$&$0.1h$    &$0.2h$    &$0.3h$    &$0.4h$    &$0.5h$    &$0.6h$   \\
            \hline
            MOTA&	79.06\%&	79.32\%&	79.27\%&	79.21\%&78.85\%&	78.65\%   \\
            IDsw&	78&	79&	72&	77&	94&	114   \\
            \bhline
    \end{tabular}
    }
\end{table}

\section{Conclusion}
\label{sec:conclusion}

In this paper, we proposed Gr-IoU, a novel approach to address data association errors in MOT.
Gr-IoU incorporates 3D constraints and transforms the coordinates of detected bounding boxes onto the ground plane, thereby computing IoU based on these transformed rectangles.
This method alleviates redundancy in the cost matrix and enhances matching efficiency.
Experiments on the MOT17 and MOT20 datasets show that Gr-IoU outperforms existing methods such as SORT and ByteTrack in terms of MOTA, IDF1, and other key metrics.

As a future work, there is room for improvement in the method for selecting the parameter $d$.
In addition, further improvements in accuracy are expected by applying camera calibration and incorporating appearance features into Gr-IoU.



%
%
\bibliographystyle{splncs04}
\bibliography{main}

\begin{thebibliography}{10}
\providecommand{\url}[1]{\texttt{#1}}
\providecommand{\urlprefix}{URL }
\providecommand{\doi}[1]{https://doi.org/#1}

\bibitem{aharon2022bot}
Aharon, N., Orfaig, R., Bobrovsky, B.Z.: Bot-sort: Robust associations multi-pedestrian tracking. arXiv preprint arXiv:2206.14651  (2022)

\bibitem{Bergmann_2019tracktor}
Bergmann, P., Meinhardt, T., Leal-Taixe, L.: Tracking without bells and whistles. In: 2019 IEEE/CVF International Conference on Computer Vision (ICCV). IEEE (Oct 2019)

\bibitem{bernardin2008evaluating}
Bernardin, K., Stiefelhagen, R.: Evaluating multiple object tracking performance: the clear mot metrics. EURASIP Journal on Image and Video Processing  \textbf{2008},  1--10 (2008)

\bibitem{bewley2016sort}
Bewley, A., Ge, Z., Ott, L., Ramos, F., Upcroft, B.: Simple online and realtime tracking. In: 2016 IEEE international conference on image processing (ICIP). pp. 3464--3468. IEEE (2016)

\bibitem{cao2023observation}
Cao, J., Pang, J., Weng, X., Khirodkar, R., Kitani, K.: Observation-centric sort: Rethinking sort for robust multi-object tracking. In: Proceedings of the IEEE/CVF conference on computer vision and pattern recognition. pp. 9686--9696 (2023)

\bibitem{chen2018motdt}
Chen, L., Ai, H., Zhuang, Z., Shang, C.: Real-time multiple people tracking with deeply learned candidate selection and person re-identification. In: 2018 IEEE international conference on multimedia and expo (ICME). pp.~1--6. IEEE (2018)

\bibitem{dendorfer2020mot20}
Dendorfer, P., Rezatofighi, H., Milan, A., Shi, J., Cremers, D., Reid, I., Roth, S., Schindler, K., Leal-Taix{\'e}, L.: Mot20: A benchmark for multi object tracking in crowded scenes. arXiv preprint arXiv:2003.09003  (2020)

\bibitem{du2023strongsort}
Du, Y., Zhao, Z., Song, Y., Zhao, Y., Su, F., Gong, T., Meng, H.: Strongsort: Make deepsort great again. IEEE Transactions on Multimedia  \textbf{25},  8725--8737 (2023)

\bibitem{fischler1981random}
Fischler, M.A., Bolles, R.C.: Random sample consensus: a paradigm for model fitting with applications to image analysis and automated cartography. Communications of the ACM  \textbf{24}(6),  381--395 (1981)

\bibitem{ge2021yolox}
Ge, Z., Liu, S., Wang, F., Li, Z., Sun, J.: Yolox: Exceeding yolo series in 2021. arXiv preprint arXiv:2107.08430  (2021)

\bibitem{kalman1960new}
Kalman, R.E.: A new approach to linear filtering and prediction problems  (1960)

\bibitem{kuhn1955hungarian}
Kuhn, H.W.: The hungarian method for the assignment problem. Naval research logistics quarterly  \textbf{2}(1-2),  83--97 (1955)

\bibitem{milan2016mot16}
Milan, A., Leal-Taix{\'e}, L., Reid, I., Roth, S., Schindler, K.: Mot16: A benchmark for multi-object tracking. arXiv preprint arXiv:1603.00831  (2016)

\bibitem{suarez2022elsed}
Su{\'a}rez, I., Buenaposada, J.M., Baumela, L.: Elsed: Enhanced line segment drawing. Pattern Recognition  \textbf{127},  108619 (2022)

\bibitem{wang2024smiletrack}
Wang, Y.H., Hsieh, J.W., Chen, P.Y., Chang, M.C., So, H.H., Li, X.: Smiletrack: Similarity learning for occlusion-aware multiple object tracking. In: Proceedings of the AAAI Conference on Artificial Intelligence. vol.~38, pp. 5740--5748 (2024)

\bibitem{wang2019towardsjde}
Wang, Z., Zheng, L., Liu, Y., Wang, S.: Towards real-time multi-object tracking. The European Conference on Computer Vision (ECCV)  (2020)

\bibitem{wojke2017deepsort}
Wojke, N., Bewley, A., Paulus, D.: Simple online and realtime tracking with a deep association metric. In: 2017 IEEE international conference on image processing (ICIP). pp. 3645--3649. IEEE (2017)

\bibitem{zhang2022bytetrack}
Zhang, Y., Sun, P., Jiang, Y., Yu, D., Weng, F., Yuan, Z., Luo, P., Liu, W., Wang, X.: Bytetrack: Multi-object tracking by associating every detection box. In: European conference on computer vision. pp. 1--21. Springer (2022)

\bibitem{zhang2021fairmot}
Zhang, Y., Wang, C., Wang, X., Zeng, W., Liu, W.: Fairmot: On the fairness of detection and re-identification in multiple object tracking. International journal of computer vision  \textbf{129},  3069--3087 (2021)

\bibitem{zhou2020tracking}
Zhou, X., Koltun, V., Kr{\"a}henb{\"u}hl, P.: Tracking objects as points. In: European conference on computer vision. pp. 474--490. Springer (2020)

\end{thebibliography}
\end{document}